\documentclass{article} %
\usepackage{iclr2025_conference,times}

\usepackage{amsmath,amsfonts,bm}

\def\eqref#1{equation~\ref{#1}}

\def\1{\bm{1}}

\DeclareMathAlphabet{\mathsfit}{\encodingdefault}{\sfdefault}{m}{sl}
\SetMathAlphabet{\mathsfit}{bold}{\encodingdefault}{\sfdefault}{bx}{n}

\usepackage{hyperref}
\usepackage{url}
\usepackage{booktabs}

\usepackage{graphicx} 

\usepackage{algorithm}
\usepackage{algorithmic}

\usepackage{amsmath}
\usepackage{amsfonts}
\usepackage{booktabs}
\usepackage{wrapfig}
\usepackage[table]{xcolor} %
\definecolor{rowgray}{RGB}{242,242,242}  %
\usepackage{xcolor}

\newcommand{\reb}[1]{\textcolor{black}{#1}}
\newcommand{\remove}[1]{}
\usepackage{multirow}
\usepackage{soul}
\usepackage[rightcaption]{sidecap} %

\usepackage{capt-of}

\title{
SLIM-Brain: A Data- and Training-Efficient Foundation Model for fMRI Data Analysis

}

\iclrfinalcopy

\author{%
	Mo Wang\textsuperscript{\rm 1,\rm 2,\thanks{Equal contribution.}},
	Junfeng Xia\textsuperscript{\rm 1,\footnotemark[1]},
	Wenhao Ye\textsuperscript{\rm 1},
	Enyu Liu\textsuperscript{\rm 1},
	Kaining Peng\textsuperscript{\rm 1},
    Jianfeng Feng\textsuperscript{\rm 3},\\
    \textbf{Quanying Liu\textsuperscript{\rm 1,\thanks{Corresponding authors.}}},
    \textbf{Hongkai Wen\textsuperscript{\rm 2,\footnotemark[2]}}\\
  \textsuperscript{\rm 1}Department of Biomedical Engineering, Southern University of Science and Technology, China\\
  \textsuperscript{\rm 2}Department of Computer Science, University of Warwick, The UK\\
	\textsuperscript{\rm 3}Institute of Science and Technology for Brain-Inspired Intelligence, Fudan University, China
    \\
	\texttt{liuqy@sustech.edu.cn; hongkai.wen@warwick.ac.uk}
}

\begin{document}

\maketitle

\begin{abstract}

Foundation models are emerging as a powerful paradigm for fMRI analysis, but current approaches face a dual bottleneck of data- and training-efficiency. Atlas-based methods aggregate voxel signals into fixed regions of interest, reducing data dimensionality but discarding fine-grained spatial details, and requiring extremely large cohorts to train effectively as general-purpose foundation models. Atlas-free methods, on the other hand, operate directly on voxel-level information - preserving spatial fidelity but are prohibitively memory- and compute-intensive, making large-scale pre-training infeasible. 
We introduce \textbf{SLIM-Brain} (\textbf{S}ample-efficient, \textbf{L}ow-memory fMR\textbf{I} Foundation \textbf{M}odel for Human \textbf{Brain}), a new atlas-free foundation model that simultaneously improves both data- and training-efficiency. SLIM-Brain adopts a two-stage adaptive design: (i) a lightweight temporal extractor captures global context across full sequences and ranks data windows by saliency, and (ii) a 4D hierarchical encoder (Hiera-JEPA) learns fine-grained voxel-level representations only from the top-$k$ selected windows, while deleting about 70\% masked patches.
\reb{Extensive experiments across seven public benchmarks show that SLIM-Brain establishes new state-of-the-art performance on diverse tasks, while requiring only 4 thousand pre-training sessions and approximately 30\% of GPU memory comparing to traditional voxel-level methods.Code and trained weights of SLIM-Brain are available at \href{https://github.com/OneMore1/SLIM-Brain2026}{link}.}
\end{abstract}

\section{Introduction}

\begin{wrapfigure}[18]{r}{0.41\textwidth}
 \vspace*{-2\baselineskip}  
  \centering
    \includegraphics[width=0.40\columnwidth]{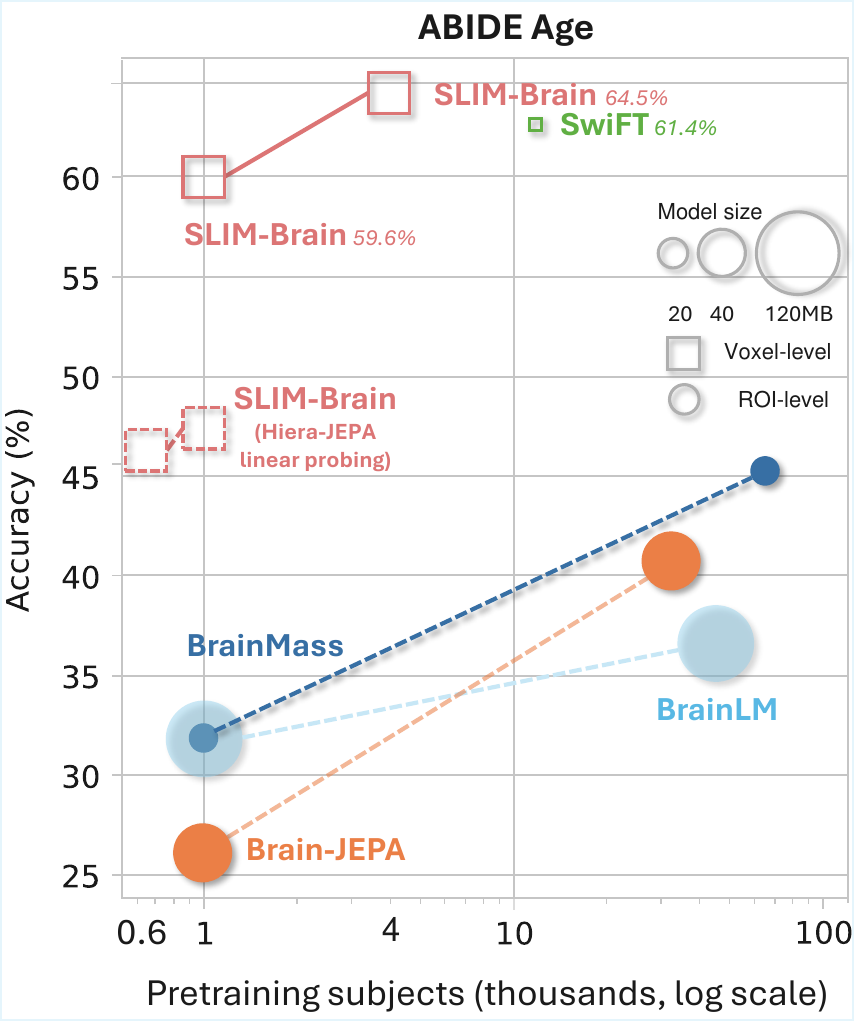}
  \caption{\textbf{Performance \& Pretraining size.}
    \reb{Our method \textit{SLIM-Brain} reaches 64.5\% age-classification accuracy with only about \textbf{4 thousand} sessions in pretraining.}
}

  \label{fig:pareto}
\end{wrapfigure}

Functional Magnetic Resonance Imaging (fMRI) has been the de facto modality for non-invasive analysis of human brain activities, with broad applications from clinical diagnostics, to monitoring neurological conditions and understanding cognitive processes~\citep{song2008brain,horikawa2017generic}. Modern MRI scanners capture brain activity via monitoring the Blood Oxygenation Level Dependent (BOLD) signals at voxel level of human brains, and are capable of acquiring high-resolution volumetric data (e.g. up-to 1$mm$ spatial resolution) over time. As a result, a single scan can generate a massive four-dimensional (4D) data sequence (3D space$\times$time), posing significant challenges to extract meaningful representations of brain activities, functional connectivity, and their associations with behavior and diseases.

\begin{figure}[t] 
    \centering
    \includegraphics[width=\textwidth]{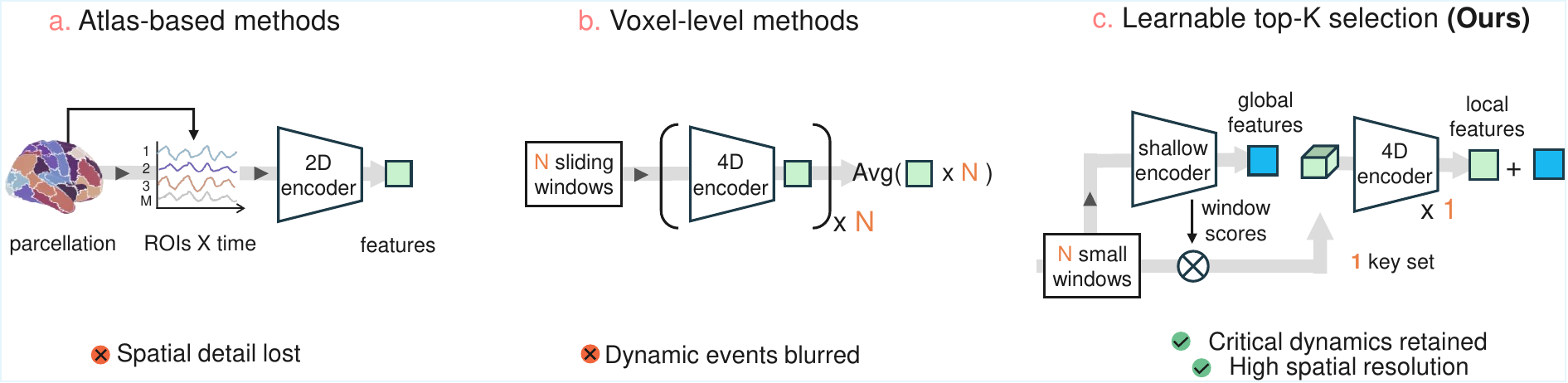}
     \vspace*{-1\baselineskip} 
\caption{
\textbf{(a) ROI-based.} Atlas parcellation coarsely downsamples space, introducing atlas bias and erasing voxel-level detail. 
\reb{\textbf{(b) Volume sliding-window pipelines.} Resolution is retained, but fixed window lengths (e.g., 40 frames) with simple averaging dilute transient events and miss cross-window dynamics. 
\textbf{(c) Ours.} A lightweight global pass ranks windows; the top small windows  (e.g., 5 frames) are concatenated to a set (e.g., 40 frames) and encoded with a 4D encoder and fused with global features, yielding efficient multi-granularity} representations with fine spatial semantics and long-range spatiotemporal structure. \vspace*{-1\baselineskip} }
\label{fig:topk}
\end{figure}

Instead of directly processing the massive volumetric fMRI data, existing studies often rely on atlas-based parcellations, where voxel-level signals are aggregated within template-defined anatomical brain regions - referred to as Regions of Interest (ROIs) in the following -  according to an ``atlas'', effectively converting the 4D data into a lower dimensional 2D format: i.e., ROIs \(\times\)time  (Fig.~\ref{fig:topk}a). They then apply signal processing and/or machine learning models tailored for specific tasks on such 2D data, e.g. disease classification. However, these supervised approaches typically require labeled data to train their models, while recent studies~\citep{Marek2022ReproducibleBA} have shown that to achieve statistically reliable results, very large cohorts (e.g. often $>$1000 participants) are necessary - adding yet another layer of challenges in practical fMRI analysis.

More recently, there has been a growing interest in developing deep learning based \textit{foundation models} for fMRI data analysis, inspired by their remarkable performance in Computer Vision and Natural Language Processing tasks. The idea is to pre-train (usually in a self-supervised way) a large model to learn general-purpose representations of brain activity on vast neuroimaging data, which can then be adapted to diverse downstream tasks with limited labeled data (e.g., via fine-tuning). Broadly speaking, current efforts on building foundation models, or more generally applying deep learning techniques in fMRI data analysis can be categorized into two threads: \textit{atlas-based} and \textit{atlas-free} approaches. The former continues the traditional paradigm of summarizing voxel-level signals into predefined brain regions~\citep{caro2023brainlm, assran2023self, yang2024brainmass}, therefore leveraging anatomical priors and improving interpretability. However those methods bear several key limitations: i) there is no universally optimal atlas: performance on different downstream tasks may highly depend on parcellation choices, where results across studies using heterogeneous atlas are not directly comparable~\citep{wang2025dca,salehi2020there}; ii) averaging based on any pre-defined atlas will inevitably discard important voxel-level information - as a result such models often need to be trained on very large cohorts (e.g., $\sim60k$ \reb{sessions}) to perform well~\citep{dong2024brain,caro2023brainlm} (see Fig.~\ref{fig:pareto}); and iii) any analysis using such models will be confined to the parcel resolution of the atlas used, where probing within-region structures (e.g., isolating amygdalar subnuclei during fear conditioning) is impossible~\citep{Wen2022TemporallyAA}.

On the other hand, atlas-free methods~\citep{zhao2018modeling,vu2020fmri,nguyen2020attend} aim to learn directly from voxel-level data without imposing ROI boundaries of certain atlas, allowing the learned models to capture fine-grained functional patterns and potentially discover novel brain organization. However, existing atlas-free approaches have typically been developed for specific tasks rather than as general-purpose foundation models, due to their prohibitively high training cost. The deep learning architectures underlying those models, such as the widely adopted Vision Transformers (ViT), incur quadratic cost (both memory and compute) with respect to input dimensions. This makes it nearly impossible to pre-train them on large-scale fMRI data, and thus building atlas-free foundation models still remains an open challenge. 
Recent work has explored efficient variants such as Shifted-window (Swin) Transformer~\citep{kim2023swift, Sun2025VoxelLevelBS,Kwon2024PredictingTB,peng2025whole}, but at each timestamp they still feed the entire dense fMRI volume into the encoder, wasting resources on those voxels ($\approx$70\%) outside the brain with no valid signal. Similar inefficiencies also arise along the temporal axis, where existing approaches either only train with data pertain to specific task states, e.g., extracting $\sim$30 timestamps~\citep{shi2023self,Sun2025VoxelLevelBS,rosenman2024pre}, or consider sliding windows with limited sizes, where data within each window is processed independently through the model, with results aggregated thereafter~\citep{kim2023swift}(e.g., as shown in Fig.~\ref{fig:topk}b). In practice neither is ideal: the former obviously hurts generalization capabilities of the model beyond task states, while the latter evenly processes every window - may lose focus on those truly important data segments - leading to inferior performance comparing to some of the recent atlas-based approaches in out-of-distribution cases~\citep{dong2024brain}.

To address these challenges, in this paper we propose \textbf{SLIM-Brain}, a new atlas-free foundation model for fMRI analysis that overcomes the limitations of both existing atlas-based and atlas-free approaches. SLIM-Brain achieves the best of both worlds by jointly pushing data and training efficiency: it requires much less data during pre-training - outperforming state-of-the-art atlas-based models with only a fraction of their training data, while at the same time reducing memory/compute by an order of magnitude compared to the most recent atlas-free methods. At its core, SLIM-Brain adopts a novel two-stage adaptive paradigm, working in tandem across the temporal and spatial domains: i) a lightweight temporal extractor that performs coarse sweeps over full sequences, capturing global context and identifying the top-$k$ most informative data windows as shown in Fig.~\ref{fig:topk}c; and ii) an efficient encoder based on hierarchical Joint Embedding Predictive Architecture (Hiera-JEPA) that delves into the selected windows, but only focusing on voxels with valid signals rather than processing the full volumes. Concretely, the technical contributions of this paper are as follows:

\begin{itemize}
  \item To the best of our knowledge, we are the first to systematically study the dual challenges of data efficiency (the heavy reliance on extremely large cohorts) and training efficiency (prohibitively high memory and compute costs) in foundation models for fMRI analysis, arguing that these are the primary obstacles preventing the widespread adoption of such models in practical settings.
  \item We propose SLIM-Brain, a data- and training-efficient foundation model built upon a two-stage adaptive pipeline. First, a lightweight extractor captures coarse global context and ranks \reb{small temporal windows. Next, only the selected data windows are concatenated and processed} by a hierarchical Joint Embedding Predictive Architecture (Hiera-JEPA) encoder, which focuses exclusively on voxels with valid signals, discarding approximately 70\% masked patches. This adaptive design yields fine-grained voxel-level features while reducing GPU memory usage to about 30\% of Swin-based models. \remove{Finally, we fuse global context with voxel-level features to produce an end-to-end representation of the fMRI sequence.}
  \item  We validate SLIM-Brain extensively across multiple downstream tasks (e.g. sex, age and fingerprint classification) and four different datasets. Results show that SLIM-Brain establishes the new state-of-the-art performance on diverse tasks, while requiring only a tiny amount of data 
  (1$k$ vs. 32$k$ \reb{sessions}) 
  in pre-training 
  than the strongest baselines. 
\end{itemize}

\section{Related Work}

\noindent\textbf{Foundation Models for fMRI Analysis.}
fMRI signals reflect ongoing brain states and cognitive processes. Early work framed decoding as supervised classification or regression on activity patterns, which often produced task-specific features with limited out-of-distribution generalization~\citep{song2008brain,horikawa2017generic,ye2023explainable,zhao2018modeling,vu2020fmri,nguyen2020attend}. More recently, the field has pivoted to task-agnostic foundation models trained with self-supervised pipelines on unlabeled data, aiming for representations that transfer across tasks and datasets.
One prominent line of work pools voxel signals within a predefined atlas to obtain region-wise time series and then learns objectives on those ROI tokens—either generative (masked reconstruction, as in BrainLM) or latent prediction (as in Brain-JEPA)~\citep{caro2023brainlm,dong2024brain}. These approaches are memory-efficient and scalable, but their quality can depend on the chosen parcellation and its granularity. A related direction converts the atlas-reduced time series into pairwise or higher-order functional graphs and optimizes unsupervised objectives on these structures~\citep{yang2024brainmass,thapaliya2024dsamad,han2025hypergraph}. For example, BrainMass augments networks by randomly dropping time points from the BOLD signal during training to encourage robustness~\citep{yang2024brainmass}, and hypergraph formulations have been explored to capture higher-order relationships~\citep{han2025hypergraph}.
Complementing these atlas-based routes, volumetric encoders operate directly on 4-D fMRI volumes, thereby avoiding atlas-induced bias and preserving fine spatial detail~\citep{Peng2025WholebrainTR,Sun2025VoxelLevelBS,Kwon2024PredictingTB,kim2023swift}.In practice, computational and memory constraints often necessitate training on short, task-aligned excerpts rather than full-length recordings. Typical applications include task classification \citep{shi2023self}, brain-state decoding \citep{Sun2025VoxelLevelBS}, and stress prediction \citep{rosenman2024pre}. Turning these systems into a general-purpose foundation model remains a major challenge.

\noindent\textbf{Efficient Vision Transformers.}
Vanilla Vision Transformers (ViTs) apply global self-attention over all tokens. The cost scales quadratically with token count and the model offers weak spatial inductive bias~\citep{dosovitskiy2020image}. This is impractical for high-resolution or long 4-D fMRI sequences, where a single scan can yield millions of voxel tokens across time.
Hierarchical Transformers address these issues by progressively downsampling tokens and widening channels (e.g., MViT, Swin)~\citep{fan2021multiscale,liu2021swin}. For example, Swin restricts attention to local windows and cyclically shifts the partition between layers, which stabilizes memory and accuracy on dense inputs. However, it limits global context to multi-layer message passing, ties the model to a dense, regular lattice, and introduces engineering overhead that complicates masked-token pretraining. These drawbacks are magnified in fMRI: spatial grids are high resolution and a large fraction of voxels are non-brain background.
Hiera shows that many hand-crafted components are unnecessary. A minimal hierarchical ViT by pretext task achieves superior speed/accuracy trade-offs and is naturally compatible with MAE-style pretraining~\citep{ryali2023hiera}. For voxel-level 4-D fMRI, this is particularly attractive: with MAE-style masking, non-informative background can be excluded from the encoder, reducing GPU memory and improving throughput while preserving fine-grained brain signals.

\begin{figure*}[h] 
    \centering
    \includegraphics[width=\textwidth]{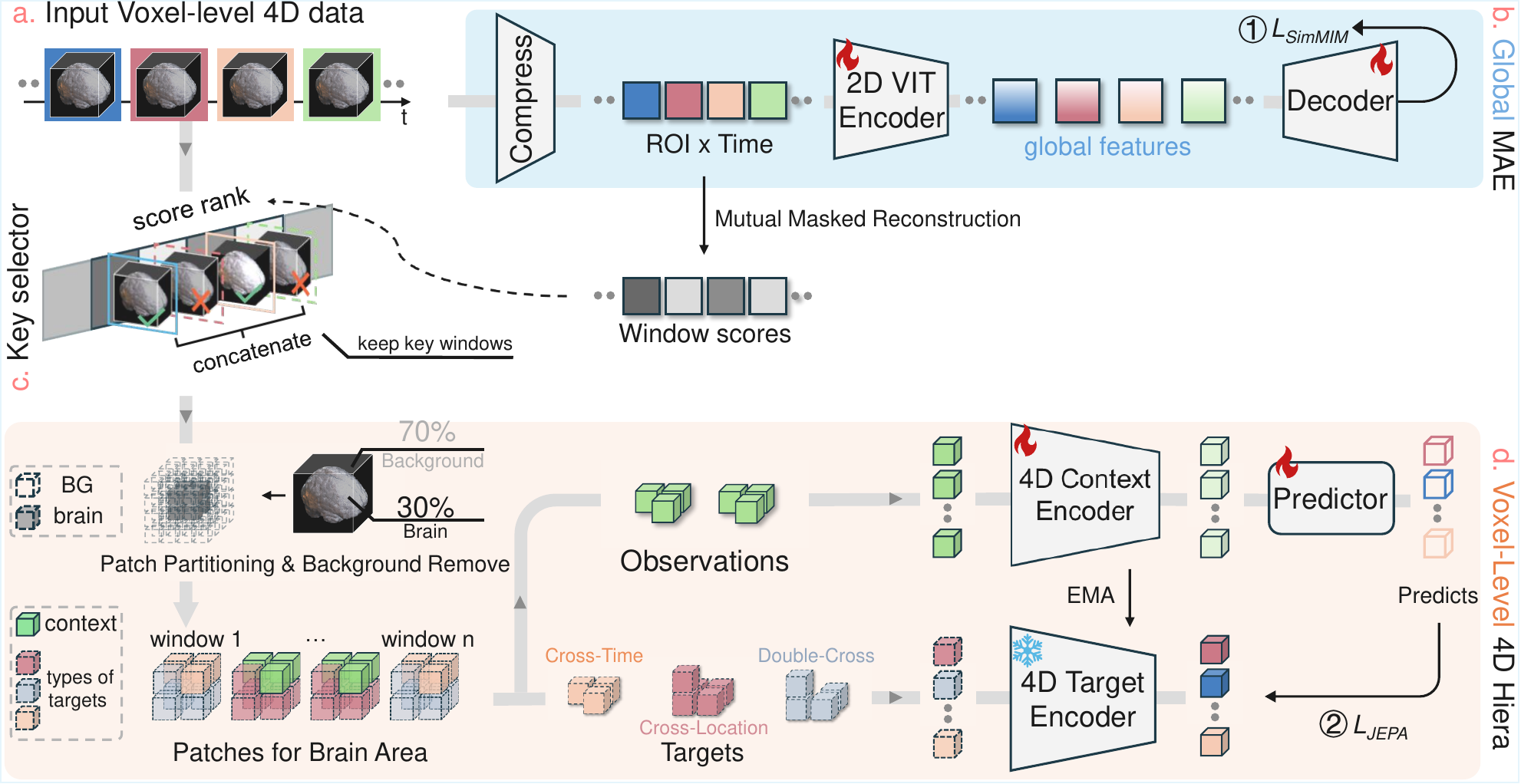}
     \vspace*{-1\baselineskip} 
    \caption{
    \textbf{SLIM-Brain Pipeline.}
    (a) Atlas-free 4D fMRI input at voxel resolution. 
    (b) A lightweight ViT  processes the full recording to produce robust global features. 
    (c) Using the same masking mechanism, a cross-window masked-reconstruction score ranks temporal windows and selects informative segments. 
    (d) Selected windows are routed to a voxel-level 4D Hiera encoder to extract fine-grained  representations without any predefined atlas. 
    \remove{The final fMRI representation is obtained by fusing global features and fine-grained key-window features.}}
    \label{fig:pipeline}
\end{figure*}

\section{Methods}

SLIM-Brain is an atlas-free 4D fMRI encoder that operates directly on voxel-level volumes (Fig.~\ref{fig:pipeline}a).
We first summarize full-length recordings with a lightweight ViT trained using masked autoencoding (MAE; SimMIM-style) to obtain robust global features (Fig.~\ref{fig:pipeline}b).
Using the same masking machinery, we compute a \emph{mutual masked reconstruction} score to rank temporal \emph{windows} and select the informative ones (Fig.~\ref{fig:pipeline}c).
The selected windows are then routed to a 4D Hiera encoder, which extracts fine-grained voxel-level representations without any predefined atlas (Fig.~\ref{fig:pipeline}d).
This selective-compute design avoids pushing large non-brain background through the encoder and sidesteps redundant volumetric reconstruction, yielding substantial gains in speed and GPU memory efficiency while preserving voxel-level detail.

\subsection{Global features via masked autoencoding}

\reb{Given a 4D fMRI time series
$\mathbf{X}\in\mathbb{R}^{H\times W\times D\times T}$,
we first partition the spatial volume $(H,W,D)$ into non-overlapping cubic patches of size $(u,u,u)$, yielding $B$ candidate patches. Using a brain mask, we discard any patch whose voxels are all background, resulting in $b$ valid patches. For each valid patch, we average the voxel values within the patch at every time point to obtain a single time series of length $T$. Stacking these patch-wise time series yields a 2D matrix
$\mathbf{X}'\in\mathbb{R}^{b\times T}$,
where the first dimension $b$ indexes the retained patches and the second dimension $T$ indexes time.}

Next, we partition the temporal axis of $\mathbf{X}'$ into non-overlapping patches of length $p$ (zero-padding the tail if needed), producing
\[
\mathbf{X}''\in\mathbb{R}^{b\times M\times p}, 
\qquad 
M=\left\lceil \tfrac{T}{p}\right\rceil .
\]
By stacking the $bM$ patches, we form a token matrix 
$\mathbf{P}\in\mathbb{R}^{(bM)\times p}$.

The global branch follows a masked autoencoding (MAE) scheme ~(Fig.~\ref{fig:pipeline}b). 
We randomly mask a ratio $r$ of tokens and feed the whole masked data $\mathbf{P}_{\text{masked}}$ to a Vision Transformer encoder $\mathcal{E}$:
\[
\mathbf{z} \;=\; \mathcal{E}\!\big(\mathbf{P}_{\text{masked}}\big)\in\mathbb{R}^{bM\times C},
\qquad
\]

where $\mathbf{z}$ is the representation. 
A lightweight reconstruction head $\mathcal{R}$ predicts the original data,
\[
\widehat{\mathbf{P}} \;=\; \mathcal{R}(\mathbf{z}) \;\in\; \mathbb{R}^{bM\times p},
\]
and training uses a SimMIM-style objective on all positions:
\[
\mathcal{L}_{\text{SimMIM}}
\;=\;
\frac{1}{bMp}\,\big\|
\widehat{\mathbf{P}}-\mathbf{P}
\big\|_2^2 .
\]

This global MAE path yields an integrated representation of the full-length sequence, preserving long-range dynamics even when only a small subset of top-ranked windows is subsequently processed by the heavy 4D encoder.

\subsection{Top-$k$ selector via mutual masked reconstruction.}
We then rank temporal windows using the pretrained frozen global MAE as a context learner \reb{which was pretrained on the same dataset as the voxel-level 4D Hiera encoder. }(Fig.~\ref{fig:pipeline}c). 
Given $M$ non-overlapping temporal windows, the MAE captures window-level structure and provides a principled signal for assessing each window’s contribution to the whole.

For a candidate window $m\in\{1,\dots,M\}$, we keep only window $m$ and mask the remaining $M{-}1$ windows \reb{which window $\zeta$}, then run the frozen MAE to reconstruct the masked ones. 
Let $Y_j$ denote the ground-truth content of window $j$ and $\hat{Y}_j^{(m)}$ the reconstruction when only window $m$ is provided. 
We define the \emph{mutual masked reconstruction} score as the negative reconstruction error averaged over all masked windows:
\[
s_m \;=\; -\,\frac{1}{M-1}\sum_{j\neq m}\mathrm{MSE}\!\left(\hat{Y}_j^{(m)},\,Y_j\right),
\]
so that higher $s_m$ indicates stronger global representativeness (i.e., patch $m$ better supports reconstructing the rest of the sequence). 
We then select the top-$k$ windows \reb{where each window consists of $\zeta$ consecutive frames},
\[
\mathcal{T} \;=\; \operatorname{Topk}\big(\{s_m\}_{m=1}^M,\;k\big),
\]
and feed their corresponding 4D sub-volumes to the voxel-level 4D encoder to extract fine-grained spatiotemporal representations for downstream modeling. 

\subsection{Voxel-level features with a 4D Hiera encoder}
\label{subsec:hiera}

We adopt a dual-branch Hiera encoder in the voxel-level path (Fig.~\ref{fig:pipeline}d). 
Hiera is a hierarchical Transformer in which local self-attention is restricted to mask units, replacing the Swin-style shifted windows. 
This design accommodates irregular inputs by operating at the unit level, allowing background units and any units designated for masking to be pruned outright.
For each of the top-$k$ windows from a continuous $T$-frame fMRI sequence, we partition the volume into a regular grid of mask units of size $u$, remove units that are entirely background, and retain a sparse set of foreground units $\mathcal{U}$. 
Within each retained unit, we apply a $n{\times}n{\times}n$ patch-merge operation, and we add spatial and temporal positional encodings to preserve the full 4D structure.

The path is trained with a JEPA objective and comprises a context encoder $\mathrm{Enc}_c$ and a target encoder $\mathrm{Enc}_t$ . 
At each iteration, the context view $\mathcal{C}\subset\mathcal{U}$ is formed by sampling a spatiotemporally contiguous block covering approximately $40\%$ of units. The target view $\mathcal{M}$ is sampled to be non-overlapping with the context and instantiated in three ways:
(i) different spatial units at the same time, 
(ii) the same spatial units at different times, and 
(iii) a spatiotemporal combination of (i) and (ii). 
Crucially, only the units that belong to a given view are fed to the corresponding encoder—there is no need to feed all patches and then mask them as in Swin-style tiling.

We encode the two views with a dual-branch Hiera:
\[
H_c \;=\; \mathrm{Enc}_c\big(X_{\mathcal{C}}\big), 
\qquad 
H_t \;=\; \mathrm{Enc}_t\big(X_{\mathcal{M}}\big)\ \text{(stop-grad)} .
\]
A ViT-style predictor $\mathrm{Pred}$ maps context features to the target space at the masked indices,
producing $\widehat{H}_t \;=\; \mathrm{Pred}(H_c;\mathcal{M}) \in \mathbb{R}^{|\mathcal{M}|\times C}$.
We optimize a JEPA-style regression with Smooth-$\ell_1$:
\[
\mathcal{L}_{\mathrm{JEPA}}
\;=\;
\frac{1}{|\mathcal{M}|}\sum_{p\in\mathcal{M}}
\mathrm{SmoothL1}\!\big(\widehat{H}_t[p],\, H_t[p]\big).
\]
The target branch is an exponential–moving–average (EMA) copy of the context branch:
\[
\theta_t \leftarrow \tau\,\theta_t + (1-\tau)\,\theta_c,\quad \tau\in[0,1),
\]
where $\theta_c$ and $\theta_t$ are the parameters of $\mathrm{Enc}_c$ and $\mathrm{Enc}_t$.

\subsection{Inference pipeline.}
Given a full-length fMRI sequence, the global MAE branch yields a compact global descriptor $\mathbf{z}\in\mathbb{R}^{C}$ and ranks temporal windows via mutual masked reconstruction. 
\reb{The top-$k$ windows are concatenated and passed to the 4D Hiera encoder,} whose outputs are subsequently pooled to yield a fine-grained local descriptor 
$g_{\text{top-k}} \in \mathbb{R}^{C_{\text{mid}}}$. 
\remove{The final feature for downstream modeling is obtained by channel-wise concatenation of the global vector and the pooled local descriptor:
\[
h = \text{Concat}\big(z, \text{Pool}(g_{\text{top-k}})\big) \in \mathbb{R}^{C + C_{\text{mid}}}.
\]}
We then apply average pooling followed by simple MLP layers for downstream tasks. For linear probing, the backbone is frozen and only the MLP is trained, whereas for fine-tuning, the entire model is updated end-to-end.

\section{Experiment}

\subsection{Dataset}
We leverage self-supervised pretraining on four public neuroimaging datasets: a single HCP session \citep{van2013wu}, CHCP \citep{ge2023increasing}, and two releases from the Amsterdam Open MRI Collection (AOMIC)—PIOP1 and PIOP2 \citep{snoek2021amsterdam}; the Adolescent Brain Cognitive Development (ABCD) Study~\cite{casey2018adolescent}. 
\reb{We use 70\% of the data (4129 sessions) for training, with the remainder reserved for validation (10\%) and testing (20\%).}
\reb{External validation spans four datasets: the Autism Brain Imaging Data Exchange (ABIDE) \citep{di2014autism};  the ADHD-200 Sample (ADHD) \citep{adhd2012adhd}; Alzheimer’s Disease Neuroimaging Initiative (ADNI)~\cite{jack2008alzheimer};
Parkinson’s Progression Markers Initiative (PPMI)~\cite{marek2011parkinson}.
See Appendix ~\ref{app:preproc} for preprocessing details.}

\begin{table*}[h]
\centering \Large
\renewcommand{\arraystretch}{1.3}
\caption{\reb{\textbf{Performance on external tasks.} We report the fine-tuning disease classification accuracy (\%) for \textbf{ADHD}, \textbf{ADNI} and \textbf{PPMI}; age classification and regression (ACC\% and MSE) for \textbf{ABIDE}. ``Samples (K)'' denotes the number of pre-training sessions (in thousands). The symbol * indicates statistical significance over all baselines with $p < 0.05$. Data is presented as mean $\pm$ standard deviation. }}
\resizebox{\textwidth}{!}{
\begin{tabular}{l cc c cc c}
\toprule
\multirow{2}{*}{\textbf{Model}} & \multirow{2}{*}{Samples (K)} & \multicolumn{1}{c}{\textbf{ADHD-200}} & \multicolumn{1}{c}{\textbf{ADNI (MCI)}} & \multicolumn{1}{c}{\textbf{PPMI}} & \multicolumn{2}{c}{\textbf{ABIDE age}} \\
\cmidrule(lr){3-3}\cmidrule(lr){4-4}\cmidrule(lr){5-5}\cmidrule(lr){6-7}
& & ACC\% $\uparrow$ & ACC\% $\uparrow$ & ACC\% $\uparrow$ & ACC\% $\uparrow$ & MSE $\downarrow$ \\
\midrule
BrainNetCNN & - & $54.46 \pm 2.47$ & $59.74 \pm 4.50$ & $64.24 \pm 2.41$ & $41.52 \pm 3.49$ & $0.7025 \pm 0.028$ \\
BrainGNN & - & $55.87 \pm 3.88$ & $63.20 \pm 7.38$ & $55.56 \pm 4.81$ & $33.17 \pm 4.46$ & $0.9338 \pm 0.000$ \\
BrainLM & 42 & $57.86 \pm 0.00$ & $61.41 \pm 0.09$ & $66.67 \pm 1.04$ & $39.24 \pm 4.36$ & $0.8700 \pm 0.059$ \\
BrainMass & 65 & $60.78 \pm 0.49$ & $62.39 \pm 0.60$ & $63.51 \pm 0.49$ & $48.19 \pm 1.64$ & $0.5129 \pm 0.042$ \\
Brain-JEPA & 32 & $59.74 \pm 0.23$ & $64.53 \pm 0.60$ & $64.57 \pm 1.79$ & $34.00 \pm 2.14$ & $0.2704 \pm 0.044$ \\
SwiFT & 10 & $60.81 \pm 2.38$ & $64.45 \pm 1.69$ & $58.10 \pm 0.00$ & $62.22 \pm 0.55$ & $0.4137 \pm 0.033$ \\
NeuroSTORM & 58 & $62.35 \pm 0.90$ & $66.67 \pm 1.06$ & $69.12 \pm 0.99$ & $38.64 \pm 2.14$ & $0.5890 \pm 0.066$ \\
\rowcolor{rowgray}
SLIM-Brain & 4 & \textbf{\ \ 63.53 $\pm$ 0.53*} & \textbf{\ \ 69.12 $\pm$ 1.38*} & \textbf{70.40 $\pm$ 0.59} & \textbf{\ \ 64.41 $\pm$ 0.57*} & \textbf{\ \ 0.2175 $\pm$ 0.019*} \\
\bottomrule
\end{tabular}
}
\label{table:ood_mean}
\end{table*}

\subsection{Fine-tuning results}

\reb{We fine-tune our model and evaluate it  on out-of-distribution datasets, as summarized in Table~\ref{table:ood_mean} and Appendix~\ref{app:std}. We benchmark against seven representative fMRI models—BrainNetCNN~\citep{kawahara2017brainnetcnn}, BrainGNN~\cite{li2021braingnn}, BrainLM~\citep{caro2023brainlm}, BrainMass~\citep{yang2024brainmass},  Brain-JEPA~\citep{dong2024brain}, NeuroSTORM~\cite{li2025towards} and SwiFT~\cite{kim2023swift}—using the authors’ released codes or checkpoints (pretrained on $10$k–$64$k samples). Note that BrainMass includes these datasets in its pre-training corpus, whereas our method utilizes them strictly for fine-tuning.
We adopted the medium-sized model (45M parameters) trained on 4,129 sessions as the default configuration for the experiments.
Across all the OOD evaluations, our model consistently outperforms the baselines, indicating that it learns domain-invariant representations that transfer to unseen datasets and tasks with minimal adaptation.
Statistical method (\ref{app:std}), internal experiment results (\ref{app:internal}), linear probing results results (\ref{app:linear}),  configuration details for ours  (\ref{app:setting}) are provided in the Appendix.
}

Our approach is markedly more compute- and data-efficient. 
Whereas prior work reports training Brain-JEPA for $300$ epochs on $4\times$A100 GPUs, BrainMass for $2000$ epochs on $8\times$V100 GPUs ($\sim150$ hours) \reb{ and NeuroSTORM for $30$ epochs on $4\times$A6000 GPUs (48\,GB) ($\sim13$ days) our model (4K sessions) trains for $40$ epochs on $1\times$ A100 GPUs (80\,GB) in about 20 hours, including I/O and computing.} 
On the data side, contemporary fMRI foundation models are typically pretrained at massive scale, 
for example, BrainMass is pretrained on $26$ datasets (including HCP, UKB, and ABIDE) totaling $64{,}584$ subjects, \reb{whereas our pretraining uses $4{,}129$ sessions.}

\subsection{Ablation Study}

\vspace{-2\baselineskip}
\reb{\paragraph{Ablation on key frames selection strategy.}
To assess the effect of key-frame selection on downstream performance, we evaluate four strategies for ADHD disease classification (Table~\ref{tab:topk_val}) by \textit{linear probing}. (1) \textit{Top-$k$ (ours),} 
(2) \textit{Temporal variance,} 
(3) \textit{Uniform sampling,}
(4) \textit{Random sampling.} Strategies are detailed in Appendix~\ref{app:key_frame}. The temporal variance–based method achieves competitive performance, suggesting that selecting nonredundant frames is beneficial, although its redundancy measure is less accurate than Top-$k$ (Welch’s t-test, $p = 0.0153$). Uniform and random sampling perform substantially worse, indicating that naive temporal downsampling fails to preserve discriminative disease-related dynamics (Welch’s t-test, $p = 0.0223$). Moreover, the improvement of Top-$k$ over random sampling is statistically significant (Welch’s t-test, $p = 0.0004$), confirming that the gain is not attributable to noise.}

\begin{table}[h]
\centering
\caption{\textbf{Ablation study on key frame selection strategies (ADHD).} We report Mean and STD over 3 independent runs. The symbol * indicates statistical significance ($p < 0.05$).}
\label{tab:topk_val}
\resizebox{\textwidth}{!}{ %
\begin{tabular}{lccc}
\toprule
\textbf{Strategy} & \textbf{ACC\% (Mean $\pm$ STD)} & \textbf{F1\% (Mean $\pm$ STD)} & \textbf{Description} \\
\midrule
Random & 56.0 $\pm$ 0.6 & 56.1 $\pm$ 0.7 & No heuristic \\
Uniform Sampling & 56.7 $\pm$ 1.4 & 50.9 $\pm$ 9.9 & Fixed intervals \\
Temporal Variance & 57.2 $\pm$ 1.2 & 56.4 $\pm$ 2.2 & Correlation-based \\
\rowcolor{rowgray} \textbf{Top-K (Ours)} & \textbf{\ 61.1 $\pm$ 0.5*} & \textbf{\ 61.0 $\pm$ 0.7*} & \textbf{Learnable selector} \\
\bottomrule
\end{tabular}
}
\end{table}

\newpage
\paragraph{Scaling study}

\begin{wrapfigure}[20]{r}{0.45\textwidth}
  \centering
    \includegraphics[width=0.40\columnwidth]{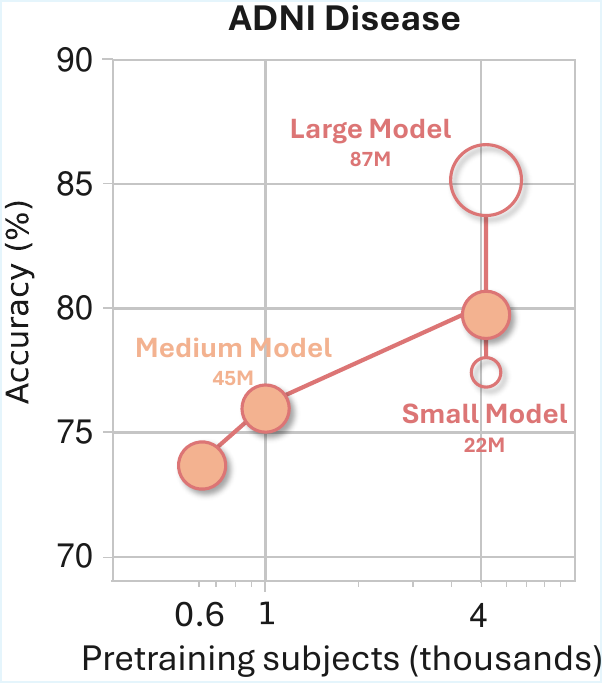}
     \vspace*{-0.7\baselineskip}  
  \caption{\textbf{Scaling study.}
      Performance on AD vs. CN at varying amounts of pre-training data and model parameter sizes..
}
\label{fig:scaling}
\end{wrapfigure}
\reb{We evaluated the scalability of SLIM-Brain by comparing its performance by ADNI (\textbf{AD} vs. CN) across varying pre-training dataset sizes (606, 1,037, 4,129) and model parameter counts (22 M, 45 M, 87 M). As illustrated in Fig.~\ref{fig:scaling}, our approach demonstrates strong scaling capabilities: accuracy improved from 73.72\% to 80.09\% as the dataset size increased, and from 77.32\% to 85.50\% as the model size increased. Crucially, we observed no signs of performance saturation even at the largest scale, suggesting that SLIM-Brain follows neural scaling laws similar to those in natural language processing. This indicates that further increasing the pre-training corpus or model capacity could yield continued gains in diagnostic precision. To ensure robustness, all experiments were conducted with three independent runs. Detailed performance metrics, including the means and variances of accuracy and F1-scores, are provided in the Appendix~\ref{app:sacling}.}

\paragraph{Top-$k$ windows outperform random selection.}

To avoid ingesting full-length fMRI sequences, we use the global MAE to obtain a coarse descriptor and to score temporal windows; only the top-ranked windows are fed to the 4D encoder (Table~\ref{tab:ablate_topk_hcp_gender}). \reb{We use five consecutive frames to form one window, and multiple such windows are grouped into a set as the input to the voxel-level model.} With a per-frame storage layout, we can load arbitrary time indices on demand, cutting $\sim80\%$ of 4D data I/O and compute when selecting the top $20\%$ of windows, comparing sliding windows average method~\citep{kim2023swift}. 
We compare \emph{Top-$k$} selection (mutual masked reconstruction) against Random selection under different frame budgets on HCP sex classification with 1K-size pre-trainging model. Top-$k$ consistently achieves higher accuracy and F1 without full-sequence 4D fMRI reads.

\begin{SCtable}[][h]
\centering
\caption{Ablation on window selection (\textbf{HCP} sex). Classification accuracy (\%) / F1-score (\%) at different frame budgets.}
\begin{tabular}{lccc}
\toprule
Set size & 5 frames (k=1) & 20 frames (k=4) & 40 frames (k=8)\\
\midrule
Random          & 83.9 / 83.7 & 84.5 / 84.1 & 86.0 / 85.9 \\
\rowcolor{rowgray}
Top\textit{-}K  & \textbf{84.5 / 84.2} & \textbf{86.7 / 86.6} & \textbf{87.7 / 87.6} \\
\bottomrule
\end{tabular}
\label{tab:ablate_topk_hcp_gender}
\end{SCtable}

\paragraph{Ablations on architectural choices.}
\reb{We assess the impact of our design decisions on memory and accuracy by 1K-size pretraining model with HCP dataset in Table.~\ref{tab:ablation:swin/jepa}. }
First, employing a 4D Hiera encoder with unit-wise attention excludes background and masked units from the forward pass, substantially reducing the memory footprint. 
Because fMRI signals are not directly interpretable like natural images, we prioritize representation quality over pixel-space fidelity and adopt a JEPA objective, dispensing with a heavy reconstruction decoder. 
With 200-frame inputs and Top-$k$ selection, our design reduces peak GPU memory from $\sim$8\,GB to $\sim$2.4\,GB \emph{while improving accuracy}.  
Meanwhile, although Hiera–MAE and Hiera–JEPA have comparable per-sample GPU memory footprints, the JEPA variant achieves substantially higher throughput.
By contrast, \citet{kim2023swift} processes ten 20-frame windows sequentially and averages their predictions; each 20-frame chunk consumes \(\approx 3.2\,\mathrm{GB}\) of GPU memory, so handling all windows inflates the per-sample memory footprint. Likewise, \citet{Kwon2024PredictingTB} employ a Swin-UNet–based model for task prediction; although it is not a foundation model and does not reconstruct to the raw signal space, it still requires \(\sim 40\,\mathrm{GB}\) of GPU memory to process 30 volumes with a batch size of four. Details are in Appendix~\ref{our_details}
\remove{Second, we ablate the roles of the global and local paths: augmenting the local 4D features with a coarse global descriptor yields a $2.5\%$ absolute gain on classification (Appendix~\ref{app:2d4d}). }

\begin{SCtable}[][h]
\centering \small
\caption{Ablation on structure choices on 1K pre-training models. \textit{Memory} denotes GPU demand per sample (200 frames; GB).}
\begin{tabular}{lcccccccc}
\toprule
\multirow{2}{*}{Model}  
& \multicolumn{2}{c}{HCP Sex} 
& \multicolumn{2}{c}{HCP Fingerprint} 
& \multicolumn{2}{c}{ABIDE Age}
& \multirow{2}{*}{memory $\downarrow$} 

\\
\cmidrule(lr){2-3}\cmidrule(lr){4-5}\cmidrule(lr){6-7}
 & ACC $\uparrow$ & F1  $\uparrow$ & ACC $\uparrow$ & F1  $\uparrow$ & ACC $\uparrow$ & F1 $\uparrow$ &  \\
\midrule
Swin-SIM     & 90.8 & 90.7 & 38.2   & 27.6   & \reb{59.3}   & \reb{59.5}  & 8.0 \\
Swin-JEPA    & 87.3 & 87.3 & 84.0 & 79.8 & \reb{\textbf{62.1}}   & \reb{\textbf{63.1}}  & 4.0 \\
Hiera-MAE    & \textbf{91.3} & \textbf{91.3} & 90.0   & 87.4   & \reb{52.6}   & \reb{52.6} & 2.4  \\ 
\rowcolor{rowgray}
Hiera-JEPA   & 91.1 & 91.1 & \textbf{98.5} & \textbf{98.1} & \reb{59.6} & \reb{58.3}  & \textbf{2.3} \\
\bottomrule
\end{tabular}
\label{tab:ablation:swin/jepa}
\end{SCtable}

\subsection{Model interpretation}

\begin{figure*}[h] 
    \centering
    \includegraphics[width=0.85\textwidth]{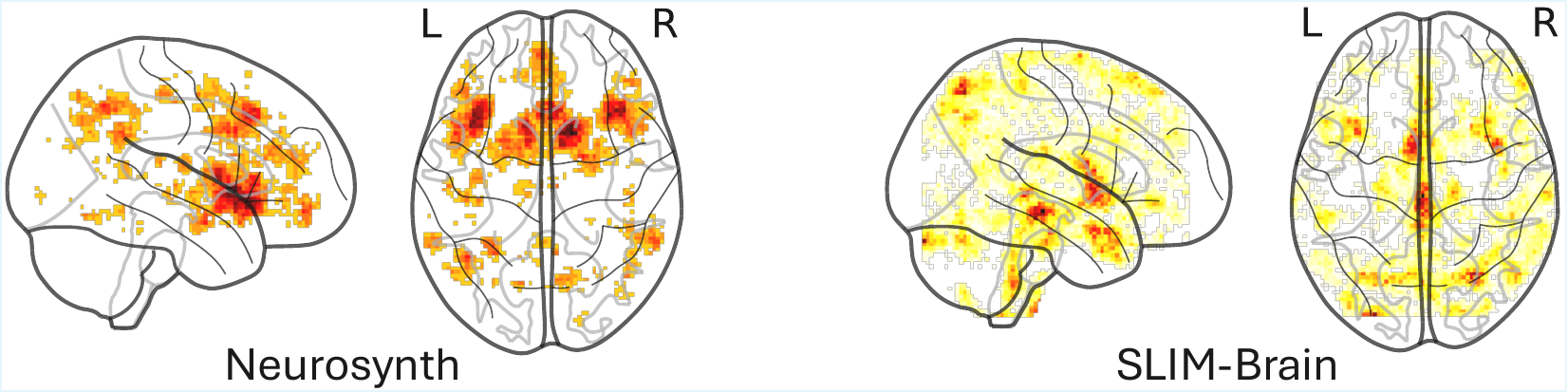}
    \caption{
    \textbf{Fine-grained attention map shows key brain area.} Left: ADHD results from Neurosynth. Right: Integrated-gradients–derived key regions from SLIM-Brain. 
}
    
    \label{fig:ig}
\end{figure*}

We examine whether SLIM-Brain’s voxel-level representations are neurobiologically meaningful. 
First, we use \emph{Neurosynth}—a large-scale, automated meta-analytic platform that aggregates findings from thousands of fMRI studies~\citep{yarkoni2011large}—to obtain disease-associated meta-analytic maps (e.g., ADHD).
Next, we derive fine-grained attention maps from the 4D Hiera encoder by projecting token-level attention back to voxel space. (Implementation details are provided in the Appendix~\ref{neurosynth}.) 
Integrated-gradients–derived key regions from SLIM-Brain substantially overlap with Neurosynth meta-analytic distributions (Fig.~\ref{fig:ig}).
The highlighted regions include dorsolateral prefrontal cortex and anterior cingulate (executive control), inferior parietal and precuneus/posterior cingulate (default mode network), as well as striatal areas, consistent with ADHD-related fronto-striatal and fronto-parietal dysfunctions.
These observations indicate that the learned features are not only predictive but also align with established neurobiological patterns, offering interpretable, voxel-level evidence directly from raw fMRI inputs.

\section{Discussion}

Recent fMRI foundation models have demonstrated strong performance on diverse downstream tasks, but most simplify inputs via \emph{atlas-based parcellation}: the brain is first divided into template-defined regions and BOLD signals are averaged within each ROI. While this streamlines computation, it can introduce parcellation bias and suppress voxel-level transients that matter for fine-grained cognitive decoding. At the other extreme, feeding full 4D volumes avoids parcellation but is computationally demanding and, when paired with global mean pooling, still risks washing out short-lived patterns.
SLIM-Brain sidesteps both spatial parcellation and coarse temporal pooling with a two-step coarse-to-fine design. 
Temporally, a top-$k$ window selector prevents full sequences from being loaded into GPU memory. 
Spatially, the Hiera--JEPA encoder prunes non-brain background patches and routes each branch to its own patch subset (context vs.\ target) without materializing the full token grid, yielding fine-grained voxel-level features while using only $\approx30\%$ of the GPU memory required by Swin-based models.
Empirically, with pretraining on $\sim$4k subjects, SLIM-Brain surpasses recent ROI/parcellation-based baselines trained on tens of thousands of subjects across multiple tasks, suggesting that voxel-level modeling captures discriminative information that atlas averaging can obscure. At the same time, the lightweight memory footprint broadens practical applicability (e.g., commodity GPUs) and makes larger-scale pretraining more tractable, pointing toward scalable, atlas-free voxel-level foundation models for fMRI.

\paragraph{Limitations and future work.}
First, although our design substantially reduces GPU memory, full-resolution 4D fMRI still imposes a nontrivial I/O burden. In practice, high-throughput storage (e.g., SSDs) and I/O-aware layouts are important. 
Second, our Top-$k$ selection uses \emph{mutual masked reconstruction}, which favors windows that best represent the remainder of the sequence. This criterion suits resting-state data but may down-weight task-evoked segments precisely because they are unique; hybrid scores that blend representativeness and event uniqueness are a natural extension. 
\remove{Third, in small-sample regimes we occasionally observe JEPA collapse (Table~\ref{tab:ft} and ~\ref{tab:adhd_abcd_lp}), consistent with prior reports \citep{mo2024connecting}; a light single-frame MAE auxiliary loss mitigates this with minor memory overhead (Appendix~\ref{app:jepa+mae}). }

\newpage
\bibliography{iclr2026_conference}
\bibliographystyle{iclr2026_conference}

\appendix
\section{Appendix}

\subsection{Preprocessing}\label{app:preproc}
HCP and CHCP data were preprocessed using the HCP minimal preprocessing pipeline \citep{glasser2013minimal}. ABCD data were processed using the ABCD-HCP pipeline. PIOP1, PIOP2 and ISYB datasets were preprocessed using fMRIPrep \citep{esteban2019fmriprep}. For ABIDE and ADHD, we adopted preprocessed datasets provided by the Preprocessed Connectomes Project (PCP) \citep{craddock2013neuro, bellec2017neuro}. 
To ensure compatibility across datasets, we first resampled all images to a uniform spatial resolution of 2 mm isotropic using cubic B-spline interpolation. For datasets with lower temporal resolution, we further interpolated the time series to a uniform sampling rate of 0.72 s TR, also using cubic B-spline interpolation. Notably, HCP, CHCP, and ABCD natively share nearly identical temporal resolution and already meet the 2 mm spatial resolution requirement after preprocessing.

\subsection{Statistical analysis of external tasks.}
\label{app:std}

\reb{To ensure the statistical reliability of our results, we conducted 3 independent runs for all Out-of-Distribution (OOD) benchmarks.And we calculates the p-values for each pair of models based on their respective accuracy values across three runs. For each model pair, the null hypothesis is that the means of the two models' accuracies are equal, and the test assesses whether there is sufficient evidence to reject this hypothesis. The p-values are stored in a matrix and the most statistically significant comparison is identified by finding the pair with the smallest p-value. The result indicates that our model statistically significant improvements ($p < 0.05$) across these diverse tasks.}

\subsection{Internal experiments}
\label{app:internal}
We fine-tune our model and evaluate it on the held-out $20\%$ split of the HCP dataset (sex classification and subject identification/fingerprinting) as summarized in Table~\ref{tab:ft}. We benchmark against two representative fMRI foundation models, BrainMass~\citep{yang2024brainmass}, and Brain-JEPA~\citep{dong2024brain}—using the authors’ released checkpoints (pretrained on $32$k–$64$k samples). For a controlled comparison under matched data budgets, we additionally retrain each baseline on the same total number of sessions as ours and report results for their best validation checkpoints (Table~\ref{tab:ft} and Fig.~\ref{fig:pareto}).

\begin{table*}[h]
\centering
\caption{Internal task. Fine-tuning accuracy (\%) and F1-score  (\%) on \textbf{HCP} Sex and fingerprint. 
“Samples (K)” is the number of pretraining \reb{sessions} (in thousands). “LP” means results from linear probing.
}
\begin{tabular}{l c cc cc  }
\toprule
\multirow{2}{*}{Model} & \multirow{2}{*}{Samples (K)} 
& \multicolumn{2}{c}{Sex} 
& \multicolumn{2}{c}{Fingerprint}   \\
\cmidrule(lr){3-4}\cmidrule(lr){5-6}
& & Acc $\uparrow$ & F1 $\uparrow$ & Acc $\uparrow$ & F1 $\uparrow$ \\
\midrule
BrainLM      & 1  & 62.4 & 61.6 & 47.0 & 41.8 \\
Brain-JEPA   & 1  & 54.0 & 35.0 & 1.0 & 0.0   \\
BrainMass    & 1  & 78.2 & 78.0 & 43.0 & 36.0   \\
BrainLM      & 42 & 74.4 & 77.7 & 51.0 & 41.8   \\
Brain-JEPA   & 32 & 87.1 & 85.4 & 57.0 & 48.9   \\
BrainMass    & 65 & 77.2 & 77.2 & 86.0 & 82.5   \\
\rowcolor{rowgray} 
SLIM-Brain (LP) & 0.6  & 90.1 & 90.0   & 89.0   & 87.2 \\
\rowcolor{rowgray} 
SLIM-Brain (LP) & 1  & 90.6 & 90.5   & 98.5   & \textbf{98.4}  \\
\rowcolor{rowgray} 
SLIM-Brain & 1  & \textbf{91.1} & \textbf{91.1} &  \textbf{98.5} &  98.1  \\
\bottomrule
\end{tabular}
\label{tab:ft}
\end{table*}

\subsection{Linear probing Results}
\label{app:linear}
\reb{To assess the intrinsic quality of the learned representations—independent of task-specific adaptation—we conduct \emph{linear probing} (LP). LP freezes the pretrained encoder and trains a lightweight linear classifier on top, providing a direct measure of how well the latent features capture task-relevant information and their linear separability \cite{he2022masked}. We evaluate LP on two downstream tasks, focusing on ADNI diagnosis (Table~\ref{tab:adhd_abcd_lp0}). Despite pretraining on a small subset of data, our linear probes transfer robustly to out-of-distribution (OOD) settings, outperforming baseline encoders and, in some cases, even their fine-tuned counterparts. At the same time, compared to MAE-based models, our method shows less performance degradation when using linear probing (2.5\% vs. 5.1\%).}

\begin{table}[htbp]
    \centering
    \caption{\textbf{Linear probing} Comparison of linear probing performance on ADNI (MCI). We report Accuracy and F1-Score over 3 independent runs and computed mean and standard deviation.}
    \label{tab:adhd_abcd_lp0}
    \begin{tabular}{lcccc}
        \toprule
        Model & Finetune Acc  & Finetune F1  & Linprobe Acc  & Linprobe F1 \\
        \midrule
        NeuroSTORM (MAE-based) & $66.67 \pm 0.60$ & $65.58 \pm 1.74$ & $61.54 \pm 0.00$ & $46.89 \pm 0.00$ \\
        \rowcolor{rowgray}
        SLIM-Brain (Ours)      & $69.12 \pm 1.38$ & $68.96 \pm 0.26$ & $66.66 \pm 1.40$ & $64.52 \pm 1.16$ \\
        \bottomrule
    \end{tabular}
\end{table}

\subsection{Experimental settings}\label{app:setting}
Unless otherwise stated, we use 4D fMRI blocks of size $96\times96\times96\times40$ $(H,W,D,T)$ ; train for 8 epochs with a global batch size of 32 on NVIDIA L40 GPUs; use Adam with learning rate $1\times10^{-3}$ for pretraining; apply an MAE masking ratio $r=0.75$; spatially downsample to a $12\times12\times12$ lattice ($H'=W'=D'=12$); \reb{To construct the 2D global data, we partition the volume into non-overlapping cubic patches with size $u = 8$, resulting in $b = 716$ foreground blocks.}; take a clip length $T=200$; set window length $p=5$ giving $M=\lceil T/p\rceil=40$ windows; keep the top-$k$ windows with $k=8$ (i.e., 40 frames when $p=5$); use mask-unit size $u=24$ voxels per side; and a patch-merge kernel $n=6$  ($6\times6\times6$). For JEPA, the context set covers $40\%$ of foreground units per iteration (non-overlapping target set is the remainder).

To ensure a fair comparison across baseline models, we conduct experiments with each baseline using three random seeds. We begin by performing a grid search over key hyperparameters, while adhering to the default settings specified in the original implementations when available. For hyperparameters not explicitly mentioned in our introduction of the baseline models, we follow the default configurations provided. Given that accuracy can be misleading on imbalanced datasets, we report weighted F1 scores as the primary evaluation metric. 

\reb{
\subsection{Details of key frame selection}\label{app:key_frame}
To assess the effect of key-frame selection on downstream performance, we evaluate four strategies for ADHD disease classification (Table~\ref{tab:topk_val}). (1) \textit{Top-$k$.} We employ the global MAE to estimate the informativeness of each window, assigning higher scores to segments with larger reconstruction errors, which indicate lower redundancy and richer signal content. 
(2) \textit{Temporal variance.} Instead of relying on model predictions, this strategy computes a frame–frame correlation matrix which measures its correlation with all other frames. Frames with the lowest mean correlation are selected, capturing temporally unique BOLD patterns that contribute nonredundant information. 
(3) \textit{Uniform sampling.} The uniform strategy samples one frame at fixed temporal intervals, providing evenly spaced coverage across the sequence without accounting for signal variation.
(4) \textit{Random sampling.} This strategy randomly selects $k$ frames from the given 4D fMRI sequence without applying any selection heuristic. We observe that Top-$k$ consistently achieves the highest classification accuracy compared with other strategies. The temporal variance–based method achieves competitive performance, suggesting that selecting nonredundant frames is beneficial, although its redundancy measure is less accurate than Top-$k$ (Welch’s t-test, $p = 0.0153$). Uniform and random sampling perform substantially worse, indicating that naive temporal downsampling fails to preserve discriminative disease-related dynamics (Welch’s t-test, $p = 0.0223$). Moreover, the improvement of Top-$k$ over random sampling is statistically significant (Welch’s t-test, $p = 0.0004$), confirming that the gain is not attributable to noise.}

\subsection{Details of scaling study}
\label{app:sacling}
\reb{We scale our model across both model size and data size. For data scaling, we pretrained a smaller model using only the HCP dataset (606 sessions). For medium-scale pre-train, we utilized HCP, CHCP, and AOMIC datasets (1037 sessions), while for large-scale pre-train, we expanded the data to include the aforementioned datasets along with the ABCD dataset (4129 sessions).
For model scaling, we experimented with three different model sizes (22 M, 45 M, 8 7M), adjusting the embedding dimension from 32 to 96 and the depth from 16 to 24.
\\\\
Following pretraining, we evaluated the performance of the models on the ADNI (AD vs. CN). Each model was run three times, each with different random seeds, while maintaining fixed hyperparameters that were chosen from one seed. The results are shown in Fig.~\ref{fig:scaling}, table~\ref{tab:app_data} and table~\ref{tab:app_model}.}

\begin{table}[h]
\centering
\caption{Data scaling on ADNI-AD}
\begin{tabular}{l ccc}
\toprule
Data Size & Samples(k) & Acc $\uparrow$ & F1 $\uparrow$ \\
\midrule
Small  & 0.6  & 73.72 $\pm$ 0.22 & 72.11 $\pm$ 1.93\\
Medium & 1  & 75.93 $\pm$ 2.10 & 75.69 $\pm$ 0.98 \\
Large  & 4  & 80.09 $\pm$ 0.87 & 79.62 $\pm$ 0.98 \\
\bottomrule
\end{tabular}
\label{tab:app_data}
\end{table}

\begin{table}[h]
\centering
\caption{Model scaling on ADNI-AD}
\begin{tabular}{l ccc}
\toprule
Model Size & Pararm (M) & Acc $\uparrow$ & F1 $\uparrow$ \\
\midrule
Small  & 22  & 77.32 $\pm$ 1.65 & 77.12 $\pm$ 1.11\\
Medium & 45  & 80.09 $\pm$ 0.87 & 79.62 $\pm$ 0.98 \\
Large  & 87  & 85.50 $\pm$ 1.16 & 85.37$\pm$ 1.11 \\
\bottomrule
\end{tabular}
\label{tab:app_model}
\end{table}

\subsection{Detailed implementations of our ablation structures}\label{our_details}

Beyond Hiera-JEPA, we also evaluated three 4D self-supervised models—Hiera-MAE, Swin-SIM, and Swin-JEPA—whose architectures are detailed below.

\paragraph{Hiera-MAE}
We adapt Hiera-MAE to 4D fMRI by following its hierarchical encoder–decoder design. Stages 1–2 use Mask-Unit Attention, while Stages 3–4 switch to Global Attention. Between stages, Q-pooling progressively downsamples the spatio-temporal tokens to form a compact latent representation. The decoder then upsamples this latent representation to reconstruct the original 4D volume, and the reconstruction error is used as the pretraining loss. Our implementation reuses the official Hiera codebase with systematic modifications to support 4D inputs and volumetric tokenization.

\begin{table}[H]
\centering
\caption{Hiera-MAE Pre-training settings (GAS: Gradient accumulation steps; BS: Batch size)}
\label{tab:pretrain_settings_wide}
\begin{tabular}{@{}ll ll@{}}
\toprule
\textbf{config} & \textbf{value} & \textbf{config} & \textbf{value} \\
\midrule
start learning rate & $5 \times 10^{-5}$ & total batch size & 4 GAS $\times$ 8 BS \\
learning rate & $1 \times 10^{-4}$ & final learning rate & $1 \times 10^{-6}$ \\
mask ratio & 0.6 &training epochs & 10 \\
weight decay & 0.05  \\
\bottomrule
\end{tabular}
\label{tab:hiera-mae}
\end{table}

\paragraph{Swin-SIM}
Complementary to Hiera-MAE, we adopt Swin-SIM with a four-stage hierarchical encoder–decoder: Swin Transformer blocks and patch merging in the encoder yield a compact latent, and a U-Net–style decoder with CNN skip connections restores resolution. We implement this by extending Swin-UNETR and treating time as channels for 4D inputs.

\begin{table}[H]
\centering
\caption{Swin-SIM Pre-training settings (same notation as Tab.~\ref{tab:hiera-mae} )}
\label{tab:pretrain_settings_wide}
\begin{tabular}{@{}ll ll@{}}
\toprule
\textbf{config} & \textbf{value} & \textbf{config} & \textbf{value} \\
\midrule
start learning rate & $5 \times 10^{-5}$ & total batch size & 4 GAS $\times$ 4 BS \\
final learning rate & $1 \times 10^{-6}$ & training epochs & 14 \\
weight decay & $1 \times 10^{-4}$ & temporal mask ratio & 0.5 \\
spatial mask ratio & 0.6 \\
\bottomrule
\end{tabular}
\end{table}

\paragraph{Swin-JEPA}
Our Swin-JEPA design combines a four-stage Swin Transformer backbone—Swin blocks within each stage and patch-merging between stages—with a JEPA objective that predicts in latent space rather than pixel space. Concretely, an online encoder processes a context crop and, together with a predictor head, produces a latent that is trained to match the latent from an EMA-updated target encoder applied to a held-out target crop from the same 4D sample. This online-to-target alignment encourages representations that are consistent across spatial/temporal views while avoiding a reconstruction decoder.

\begin{table}[H]
\centering
\caption{Swin-JEPA Pre-training settings (same notation as Tab.~\ref{tab:hiera-mae} )}
\label{tab:pretrain_settings_wide}
\begin{tabular}{@{}ll ll@{}}
\toprule
\textbf{config} & \textbf{value} & \textbf{config} & \textbf{value} \\
\midrule
start learning rate & $3 \times 10^{-5}$ & total batch size & 8 GAS $\times$ 8 BS \\
learning rate & $1 \times 10^{-3}$ & final learning rate & $1 \times 10^{-6}$ \\
training epochs & 8 & weight decay & 0.05 \\
pred\_mask\_R\_roi\_scale & (0.15, 0.3) & pred\_mask\_T\_roi\_scale & (0.2, 0.6) \\
pred\_mask\_T\_roi\_scale & (0.2, 0.6) & pred\_mask\_T\_scale & (0.0, 0.4) \\
\bottomrule
\end{tabular}
\end{table}

Across all three models we use AdamW as optimizer with a warmup–cosine learning-rate schedule. For data sampling, each sample is partitioned into non-overlapping groups of 200 frames and was uniformly selected a contiguous 40-frame clip (random start index) as the model input.

\remove{
\subsection{Ablation on SLIM-Brain components.}\label{app:2d4d}
We first analyzed the effect of architectural choices by comparing three variants: \textit{(1)SLIM-Brain (global path):} a model with only the global pathway, which aggregates coarse representations across the entire session (Fig.~\ref{fig:pipeline}b); \textit{(2)SLIM-Brain (Hiera path):} a model with only the patch-level 4D JEPA pathway, which focuses on a few salient windows (Fig.~\ref{fig:pipeline}d); and \textit{(3)SLIM-Brain (full):} a model with both global and local branches  The results in Tab.~\ref{tab:2d4d} show that while both branches contribute to performance, the combination of local-global features with alignment supervision yields the most consistent improvements, confirming the necessity of joint modeling and representation alignment.
\begin{table*}[h]
\centering
\caption{Ablation on SLIM-Brain components. Classification accuracy (\%) and F1-score on downstream tasks by small Hiera-Jepa architecture; ``Params’’ denotes trainable parameter count. }
\begin{tabular}{lcc}
\toprule
& \multicolumn{2}{c}{\textbf{Sex}} 
\\
Model & ACC $\uparrow$ & F1 $\uparrow$  \\
\midrule
SLIM-Brain (2D)     & 62.3 & 61.5 \\
SLIM-Brain (4D)     & 87.7 & 87.6   \\
\rowcolor{rowgray} 
\textbf{SLIM-Brain (full)}   & 90.0 & 90.1  \\
\bottomrule
\end{tabular}
\label{tab:2d4d}
\end{table*}}

\subsection{Neurosynth\& Integrated gradients}
\label{neurosynth}
\paragraph{Neurosynth}
The uniformity test in Neurosynth is based on a large-scale meta-analysis of thousands of fMRI studies. For a given term (e.g., “ADHD”), Neurosynth collects all studies that reported this term and then computes, at each voxel, the proportion of studies reporting activation in that location. This is statistically compared against the overall base rate of activation across the entire database, yielding a map that highlights voxels more consistently activated in studies mentioning the target term. The resulting map thus reflects consensus activation patterns associated with the disorder.

\paragraph{Implementation details of disease-associated meta-analytic maps}
We used disease-associated meta-analytic maps from Neurosynth as input references to probe whether SLIM-Brain’s voxel-level representations align with established neurobiological findings. Specifically, each meta-analytic map is passed through the pre-trained model, and we apply Integrated Gradients (IG) to compute voxel-wise attributions with respect to the model’s predictions. 

As a baseline input, we used an empty (all-zero) volume to represent the absence of activation, and interpolated from this baseline to the actual Neurosynth map. 
The implementation follows the Captum library in PyTorch, which provides efficient IG routines. 

\remove{
\subsection{MAE as an auxiliary loss}
\label{app:jepa+mae}
During our experiments and when reproducing JEPA-based models, we observed
representation collapse in small-data / small-batch regimes: both the
student and the (EMA) teacher can converge to nearly constant, uninformative
features. To mitigate this, we add a lightweight masked
autoencoding (MAE) auxiliary loss that anchors the student to a
non-degenerate signal.
\\\\
At each iteration, we uniformly sample one frame from the context view and
apply a MAE-style masked reconstruction head to predict the corresponding
teacher-frame content (see the red arrow in Fig.~\ref{fig:app:jepa+mae}).
This adds a local pixel-space objective that regularizes the student while
leaving the main JEPA objective unchanged. The total loss is
\[
\mathcal{L}_{\text{total}}
\;=\;
\mathcal{L}_{\text{JEPA}}
\;+\;
\lambda\,\mathcal{L}_{\text{MAE}},
\qquad
\mathcal{L}_{\text{MAE}}
=
\frac{1}{|\mathcal{M'}|}
\big\|
\widehat{\mathbf{x}}_{\mathcal{M'}} - \mathbf{x}_{\mathcal{M'}}
\big\|_2^2,
\]
where $\mathcal{M'}$ is the set of masked patches on the sampled frame and
$\lambda$ controls the auxiliary strength. In practice this stabilizes
training and prevents collapse with negligible overhead.
\begin{figure}[h] 
    \centering
    \includegraphics[width=1\textwidth]{figs/app_jepa_mae.pdf}
  \caption{\textbf{MAE auxiliary loss for 4D Hiera--JEPA.}
  A random frame from the context branch is masked and reconstructed by a
  lightweight head to match the teacher’s corresponding frame (red arrow),
  providing a local signal that reduces representation collapse.}
    
    \label{fig:app:jepa+mae}
\end{figure}
}

\end{document}